\documentclass[12pt]{article}
\usepackage{amsmath}
\usepackage{graphicx}
\usepackage{enumerate}
\usepackage{natbib}
\usepackage{url} 
\usepackage{bibentry}
\usepackage{subcaption}
\usepackage{diagbox}

\usepackage{graphicx}
\usepackage{subcaption}
\usepackage{caption}
\usepackage{tikz}
\usetikzlibrary{shapes, arrows}
\usepackage{multirow}
\usepackage{amssymb}
\usepackage{mathtools}
\usepackage{amsthm}
\usepackage{microtype}
\usepackage{graphicx}
\usepackage{algorithm}
\usepackage{algorithmic}
\usepackage{booktabs}
\usepackage{amsmath}
\usepackage{enumitem}
\usepackage{float}
\theoremstyle{plain}
\newtheorem{theorem}{Theorem}
\newtheorem{proposition}{Proposition}

\newtheorem{definition}{Definition}
\theoremstyle{remark}

\newtheorem{example}{Example}
\theoremstyle{empty}

\usepackage{bibentry}
\usepackage{color}

\usepackage{mathrsfs}
\usepackage{comment}

\newcommand{\blind}{1}

\setlist[enumerate]{itemsep=0pt, topsep=0pt, parsep=0pt, partopsep=0pt}

\begin{document}

\def\spacingset#1{\renewcommand{\baselinestretch}%
{#1}\small\normalsize} \spacingset{1}

\if1\blind
{
\title{\bf Optimal Transport Learning: Balancing Value Optimization and Fairness in Individualized Treatment Rules}
\author{Wenhai Cui\\
     \small{Department of Applied Mathematics, The Hong Kong Polytechnic University, Hong Kong}\\
     Xiaoting Ji\\
     \small{Department of Applied Mathematics, The Hong Kong Polytechnic University, Hong Kong}\\
    Wen Su \\
   \small{Department of Biostatistics, City University of Hong Kong, Hong Kong}\\
     Xiaodong Yan\\
   \small{Department of Mathematics and Statistics, Xi'an Jiaotong University, China}\\
      Xingqiu Zhao\\ 
    \small{Department of Applied Mathematics, The Hong Kong Polytechnic University, Hong Kong}
  }
\date{}	
\maketitle
} \fi

\if0\blind
{

  \begin{center}
{\LARGE \bf Optimal Transport Learning: Balancing Value Optimization and Fairness in Individualized Treatment Rules}
\end{center}
  \bigskip
  \bigskip
  \bigskip
  \medskip
} \fi

\bigskip

\begin{abstract}

Individualized treatment rules (ITRs) have gained significant attention due to their wide-ranging applications in fields such as precision medicine, ridesharing, and advertising recommendations. 
However, when ITRs are influenced by sensitive attributes such as race, gender, or age, they can lead to outcomes where certain groups are unfairly advantaged or disadvantaged. To address this gap, we propose a flexible approach based on the optimal transport theory, which is capable of transforming any optimal ITR into a fair ITR that ensures demographic parity. Recognizing the potential loss of value under fairness constraints, we introduce an ``improved trade-off ITR," designed to balance value optimization and fairness while accommodating varying levels of fairness through parameter adjustment. To maximize the value of the improved trade-off ITR under specific fairness levels, we propose a smoothed fairness constraint for estimating the adjustable parameter. Additionally, we establish a theoretical upper bound on the value loss for the improved trade-off ITR. We demonstrate performance of the proposed method through extensive simulation studies and application to the Next 36 entrepreneurial program dataset.
\end{abstract}

\noindent%
\vfill
\newpage
\spacingset{1.5} 

\section{Introduction}
\label{sec:intro}

The increasing accessibility of individual-level data has driven significant interest in research on individualized treatment rules (ITRs), which aim to tailor decision-making to individual characteristics by optimizing expected treatment outcomes. Over time, a variety of methods have been developed to estimate optimal ITRs. Early approaches include model-based methods, which focus on estimating the conditional mean of outcomes given individual features and treatment assignments. Examples of such methods include Q-learning \citep{watkins1992q, murphy2003optimal, qian2011performance} and A-learning \citep{robins2004optimal}. In contrast, model-free policy search approaches directly maximize the expected outcome over a class of decision functions. These include outcome-weighted learning (OWL) \citep{zhao2012estimating}, its double robust extension \citep{zhang2012robust}, residual weighted learning \citep{zhou2017residual}, matched learning \citep{wu2020matched}, and D-learning \citep{Dlearning}.


Despite these methodological advancements, there is growing concern about fairness in the application of ITRs, particularly in domains with significant societal impact, such as healthcare \citep{obermeyer2019dissecting}, hiring \citep{dastin2022amazon}, and social welfare distribution \citep{viviano2024fair}. These applications often rely on datasets that include sensitive attributes such as gender, race, and age, which may be correlated with other features, such as interview scores or diagnostic criteria. As a result, the effectiveness of treatments can be adversely affected by discrimination related to these sensitive attributes. For example, in healthcare, studies have demonstrated that Black patients may experience less favorable outcomes in medical interactions with physicians who exhibit implicit bias \citep{penner2010aversive}. Similarly, in hiring, evidence suggests that job advertisements for high-paying and senior positions are less frequently presented to women compared to men \citep{vladimirova2025fairjob}. 

However, limited research has explored disparities in treatment decisions across different sensitive attribute groups. Recent work by \cite{fang2023fairness} proposed a fairness criterion ensuring that the quantile of the potential outcome distribution induced by ITRs exceeds a specified threshold, while \cite{viviano2024fair} estimated fair policies by identifying the Pareto optimal set. Although these studies address fairness, they do not explicitly examine discrimination arising from sensitive attributes. To address this concern, we incorporate demographic parity, a fundamental fairness criterion widely used in machine learning, into the estimation of ITRs. This criterion requires algorithmic outcomes be independent of sensitive attributes \citep{agarwal2019fair,liu2022conformalized}. 
Under this fairness constraint, \cite{kim2023fair} considered the nonparametric estimation of conditional average treatment effect (CATE) that satisfies demographic parity with binary sensitive attributes.
Nevertheless, estimating the fair ITRs presents significant challenges, primarily due to the non-smooth and non-convex nature of fairness constraints. While smoothed fairness constraints can be applied for estimating ITRs, perfect demographic parity cannot be guaranteed. To the best of our knowledge, no existing study can ensure independence between ITRs and  sensitive attributes. 


Recent studies have utilized optimal transport theory for mapping data distributions into ones that are independent of sensitive attributes, achieving demographic parity \citep{johndrow2019algorithm,wang2019repairing,gordaliza2019obtaining}. Optimal transport seeks to identify the optimal transport map that minimizes the cost of moving one probability distribution into another \citep{villani2008optimal}. For instance, \cite{jiang2020wasserstein} utilized the Wasserstein-1 optimal transport map to adjust the conditional distribution of model outputs given different sensitive attributes, transforming it into a target distribution independent of these attributes. Furthermore, \cite{chzhen2020fair} established a connection between fair regression and optimal transport theory, proposing an optimal fair predictor that minimizes squared risk while adhering to demographic parity constraints. However, the use of optimal transport raises a theoretical question: whether the value loss, which is defined as the reduction in value when comparing fair ITRs to optimal ITRs, can be controlled. Additionally, a mechanism for balancing value optimality and fairness has yet to be proposed.


In this work, we employ optimal transport to obtain fair ITRs while minimizing the upper bound of potential reduction in the value function. Specifically, given different sensitive attributes, we transport the conditional distributions of the optimal decision function to their Wasserstein barycenter. This ensures that the distribution of the post-processed decision function is independent of sensitive attributes, thereby achieving demographic parity. We propose a fairness metric called Disparate Impact (DI), for evaluating the fairness levels of estimated ITRs. To balance value optimization and fairness, we introduce an ``improved trade-off ITR," which combines the fair and optimal decision functions using a self-adjusted weight function. By adopting a non-linear weight function positively correlated with CATE, we strategically reduce value loss while maintaining a specified level of fairness, as quantified by DI.
The key contributions are listed as follows:
\begin{itemize}
\item[(i)]
We proposed a generic and versatile approach that can integrate with a wide range of existing ITR approaches. This approach enables the transformation of any estimated optimal decision function into a fair version, ensuring broad applicability across diverse domains such as healthcare, hiring, and social welfare. Different from existing methods that are often tailored to specific models or settings, the proposed approach is model-free, making it a powerful tool for researchers and practitioners aiming to incorporate fairness into their decision-making processes.

\item[(ii)]
The proposed method addresses a critical challenge in fair decision-making: the non-smooth and non-convex nature of fairness constraints. By leveraging optimal transport theory, we achieve ITRs that strictly satisfy demographic parity, a fundamental fairness criterion. This is a significant advancement over existing approaches, which often rely on approximations or relaxations that fail to guarantee strict fairness. Our method ensures that the resulting ITRs are not only fair but also maintain high predictive accuracy, striking a balance that has been elusive in prior work.

\item[(iii)]
To bridge the gap between value optimization and fairness, we introduce improved trade-off ITRs, which provide a flexible and user-adjustable fairness mechanism. This innovation allows policymakers to specify their desired fairness level, and the improved trade-off ITRs dynamically adapt to meet these criteria. 
By offering a tunable fairness parameter, our method empowers stakeholders to make informed trade-offs between fairness and performance, a feature that sets it apart from rigid, one-size-fits-all solutions.

\item[(iv)]
The theoretical analysis establishes a rigorous upper bound on the reduction of the value function for improved trade-off ITRs. This bound is primarily determined by the desired fairness level and the Wasserstein distance between the distributions of the fair and optimal decision functions. This result not only provides a theoretical guarantee for the performance of our method but also offers practical insights into the trade-offs involved in achieving fairness. 
\end{itemize}


The remainder of this paper is organized as follows. In Section 2, we present a motivating example and define the concept of fair decision functions, laying the groundwork for the subsequent analysis. Section 3 provides the detailed formulation for the estimator of the fair decision function within the optimal transport framework. In Section 4, we establish the theoretical properties of the proposed estimator, including fairness guarantees and convergence rates. We demonstrate the performance of the proposed method by conducting extensive simulation studies in Section 5 and analyzing data from the Next 36 entrepreneurial training programs in Section 6. Section 7 includes some concluding remarks and discussions over future research directions, highlighting the broader implications of our approach. Detailed proofs of the theoretical results are provided in the supplementary materials.

\section{Preliminaries}
\subsection{Individualized Treatment Rules}
In the study,  we consider the framework of binary treatment, where $A \in \mathcal{A}=\{-1,1\}$ represents the treatment an individual receives.
Let $\boldsymbol{X} \in \mathcal{X}$ be a $d$-dimensional covariate vector, and $S \in \mathcal{S}$ represent sensitive attributes, such as race, gender, skin color, or religion, where $\mathcal{S}$ is a finite set with $|\mathcal{S}|$ being its cardinality.
Suppose $R$ denotes the observed outcome (larger $R$ corresponds to the better outcome).
Let $R(-1)$ and $R(1)$ be potential outcomes with treatment $-1$ and $1$, respectively.

The ITR $ \mathcal{D}(\cdot) $ is a mapping from $ \mathcal{X} \times \mathcal{S} $ to $ \mathcal{A} $.
In general, $ \mathcal{D}(\boldsymbol{x}, s) $ can be expressed as $ \operatorname{sgn}(f(\boldsymbol{x}, s)) $, where $ \operatorname{sgn}(u)=2I(u>0)-1 $, $I(\cdot)$ is an indicator function, and $ f(\boldsymbol{x}, s) $ denotes the decision function. Suppose that the propensity score function $\pi(a|\boldsymbol{x},s):={P}(A=a \mid \boldsymbol{X}=\boldsymbol{x},S=s)\geq \tau>0$ for any $a\in \mathcal{A} $ and $ (\boldsymbol{x},s)  \in \mathcal{X}\times  \mathcal{S}$.
Then, under stable unit treatment value assumption \citep{rubin2005causal}, $R=R(-1)I(A=-1)+R(1)I(A=1)$, and no unmeasured confounders assumption \citep{rubin1974estimating}, $\{R(-1), R(1)\} \perp A \mid (\boldsymbol{X}, S)$, the value function can be expressed as:
\begin{equation}\label{e1}
\mathcal{V}(f):=E[R({\operatorname{sgn}(f)})]=E\left[\frac{I(A=\operatorname{sgn}(f(\boldsymbol{X},S)))}{\pi(A\mid\boldsymbol{X},S)} R\right],
\end{equation}
 where $R(\operatorname{sgn}(f))=R(-1)I\left\{\operatorname{sgn}(f\left(\boldsymbol{X},S\right))=-1\right\}+R(1) I\{\operatorname{sgn}(f\left(\boldsymbol{X},S\right))=1\}$.
 Additionally, the optimal decision function, denoted by $f^*$, is defined as the decision function that maximizes the value function $\mathcal{V}(f)$:
$f^* = \underset{f}{\text{argmax}} \, \mathcal{V}(f)$
and the corresponding ITR, $\operatorname{sgn}(f^*)$, is referred to as the optimal ITR, denoted by $\mathcal{D}^*$.

There are many methods for estimating the optimal decision function.
Broadly, these methods can be classified into two types of approaches. One approach focuses on directly maximizing $\mathcal{V}(f)$ \citep{zhao2012estimating, zhang2012robust, liu2018augmented}.
Another approach, known as Q-learning \citep{qian2011performance}, aims to estimate the conditional outcome $E[R | \boldsymbol{X}, S, A]$ and derive the optimal ITR as $\operatorname{sgn}(E[R | \boldsymbol{x}, s, A=1] - E[R | \boldsymbol{x}, s, A=-1])$. However, conventional methods for estimating optimal decision functions often overlook sensitive attributes such as race or gender, potentially leading to biased outcomes that disproportionately impact certain groups.

\subsection{Motivation}
To demonstrate the unfairness that can arise in estimated ITRs, consider the following illustrative example.
\begin{example}\label{example11}
Consider a non-profit training organization that decides whether to admit an applicant to its free entrepreneurship training program ($A=1$) based on the applicant's characteristics $\boldsymbol{X}=\{X_1, X_2, X_{3}\}$ and the sensitive attribute $S$.
In this scenario, the reward is an innovation score assigned by a panel, modeled as:
$$R=10 + {X_3}^2 +\left\{X_1 + X_2 - 10(1 - S)I(A=1)\right\}A + \epsilon,$$ where $\epsilon$ follows a standard normal distribution, and the covariates $X_1, X_2, X_{3}$ are independently generated from a uniform distribution $\mathcal{U}[-5,5]$. The probability of admission is $P(A=1)=0.5$. Additionally, the sensitive attribute $S \in {0,1}$ follows a Bernoulli distribution $\mathcal{B}(0.5)$, where $S=1$ represents male and $S=0$ represents female.
\end{example}

\begin{figure}[h]
\centering
\begin{minipage}{0.418\textwidth}
    \centering
    \includegraphics[width=\textwidth]{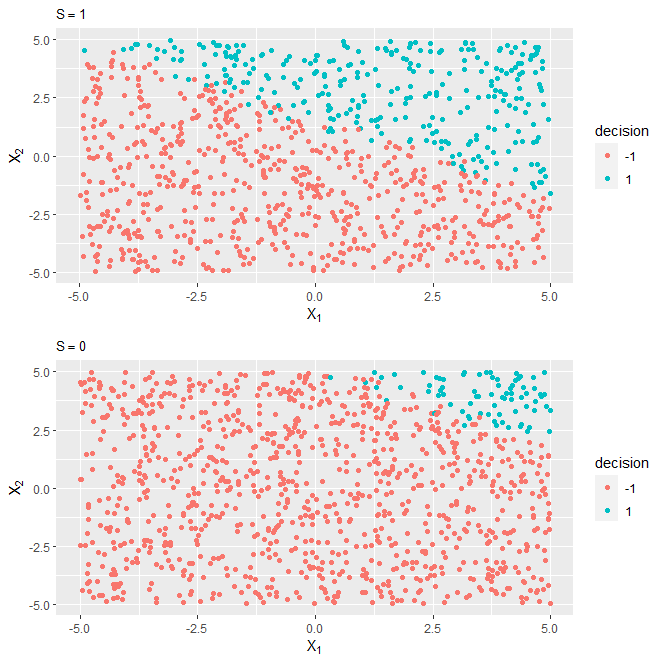}
\end{minipage}%
\begin{minipage}{0.4\textwidth}
    \centering
    \includegraphics[width=\textwidth]{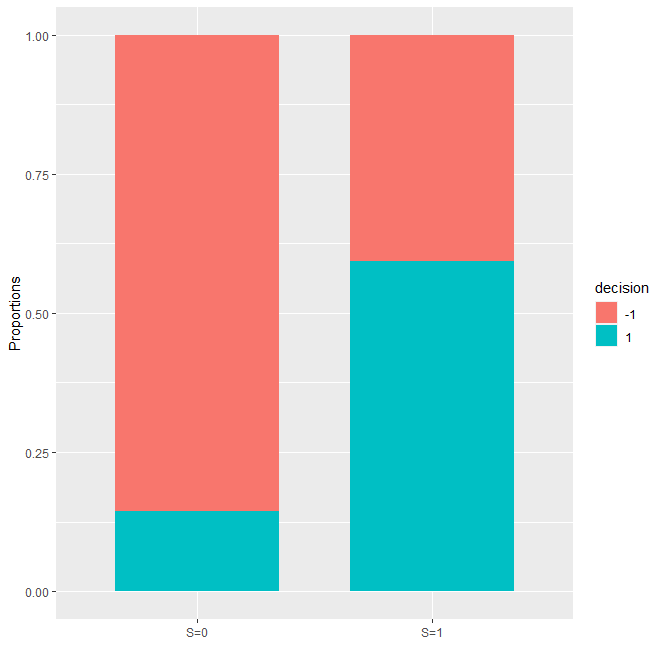}
\end{minipage}
\caption{The plot on the left illustrates the decision boundaries of the estimated optimal ITR, while the plot on the right depicts the proportions of individuals admitted to training ($A=1$) and not admitted to training ($A=-1$).
The estimated ITR exhibits significant variation based on gender (left plot), resulting in imbalanced admission proportions between different genders (right plot). The optimal ITR is estimated using the OWL method \citep{zhao2012estimating}, with both the training and testing datasets comprising 2000 samples.}
\label{fig1:side:both}
\end{figure}
Figure \ref{fig1:side:both} illustrates that the estimated ITR prioritizes males and admits a disproportionately higher number of males compared to females. Consequently, the estimated ITR, which solely aims to maximize the value function, is unfair and violates the relevant fairness laws \citep{gonzales2015fair}. Therefore, it is crucial to ensure that decision-making processes are not influenced by sensitive attributes and that welfare assignments are equitable across all subgroups.
The findings emphasize the need for methodologies that mitigate bias and
promote demographic parity in treatment decisions.
To address these concerns, we first propose a
fairness concept specifically developed for decision functions.
\begin{definition}
A decision function $f$ is defined as a fair decision function if the following condition holds:
$$
\sup_{t \in \mathbb{R}} \left| P(f(\boldsymbol{X}, S) \leq t \mid S = s) - P(f(\boldsymbol{X}, S) \leq t \mid S = s') \right| = 0,  \; s, s' \in \mathcal{S},
$$
where $\mathbb{R}$ denotes the set of real numbers. Suppose that $ \mathcal{F} = \{ f \mid f \text{ is a fair decision function} \} $ encompass all fair decision functions.
\label{def3}
\end{definition}
Let $\mathcal{D}(\boldsymbol{x}, s) = \operatorname{sgn}(g(\boldsymbol{x}, s))$. For $g \in \mathcal{F}$, we have
\begin{align*}
   &\sup _{a \in \mathcal{A}}\left|P(\mathcal{D}(\boldsymbol{X}, S) =a \mid S=s)-P\left(\mathcal{D}(\boldsymbol{X}, S) =a \mid S=s^{\prime}\right)\right|\\
   &= \left|P(g(\boldsymbol{X}, S) \leq 0 \mid S=s)-P\left(g(\boldsymbol{X}, S) \leq 0 \mid S=s^{\prime}\right)\right|\\
   &\leq \sup _{t \in \mathbb{R}}\left|P(g(\boldsymbol{X}, S) \leq t \mid S=s)-P\left(g(\boldsymbol{X}, S) \leq t \mid S=s^{\prime}\right)\right|=0.
\end{align*}
Thus, the corresponding ITR $\mathcal{D}(\boldsymbol{x}, s)$ satisfies
\begin{equation}\label{DP}
 P(\mathcal{D}(\boldsymbol{X}, S) =a \mid S=s)=P\left(\mathcal{D}(\boldsymbol{X}, S) =a \mid S=s^{\prime}\right),
\end{equation}
for any $a \in \mathcal{A}$, $s, s^{\prime} \in \mathcal{S}$. Furthermore, we regard ITRs that meet the Equation (\ref{DP}) as fair ITRs. Thus, any ITR derived from a fair decision function is a fair ITR.

\section{Methodology}
\subsection{Learning a Fair ITR}
Usually, directly estimating a decision function with fairness constraints is difficult.
To overcome this challenge, we propose a post-processing fairness algorithm that takes an unfair decision function as input and transforms it to satisfy the fairness constraint.


When we impose fairness constraints on the decision function, it inevitably results in a loss in the value function $ \mathcal{V}(f) $. This loss occurs because the fairness constraints may restrict the flexibility of the decision function. The following proposition establishes an upper bound for the value loss of any fair decision function.
\begin{proposition}
For any fair decision function $g \in \mathcal{F}$,
we have
\begin{equation}
\mathcal{V}(f^*)-\mathcal{V}(g)\leq K\sqrt{ E\left|f^*(\boldsymbol{X},S)-g(\boldsymbol{X},S)\right|^2},
\end{equation}\label{e2}
where $K=\sqrt{E[|\frac{R}{A \pi(1\mid\boldsymbol{X},S) +(1-A) / 2}|^2]}>0$ .
\end{proposition}\label{THM1}
This proposition shows that the difference between the value functions of $g$ and  $f^*$ is bounded by their $L^2$ distance.
To reduce the loss of value function, we wish to find a fair decision function $g^*$ that minimizes the upper bound of the value loss: $g^*=\arg\min_{g \in \mathcal{F}} E\left|f^*(\boldsymbol{X},S)-g(\boldsymbol{X},S)\right|^2$.
We propose to use optimal transport learning for discovering a fair decision function $g^*$ based on any optimal decision function $f^*$. 


To get $g^*$, we first introduce the Wasserstein-2 distance, which is a measure of the distance between two probability distributions.
Let $\mathcal{P}_2\left(\mathbb{R}\right)$ denote the space of Borel probability measures on $\mathbb{R}$ with finite second moments.
For probability measures $\mu, \nu \in \mathcal{P}_2\left(\mathbb{R}\right)$, $\Gamma_{\mu, \nu}$ denotes the set of all probability measures on the product space $\mathbb{R}\times \mathbb{R}$ with marginal probability measures $\mu$ and $\nu$. The Wasserstein-2 distance between $\mu$ and $\nu$  is defined as
$$ \mathcal{W}_2^2(\mu, \nu)
:=\inf _{\gamma \in \Gamma_{\mu, \nu}} \int|x-y|^2 d \gamma(x, y),x, y \in \mathbb{R}.$$


\begin{proposition}\label{thm2}
  Assume that, for each $s \in \mathcal{S}$, the conditional distribution of $f^*(\boldsymbol{X}, S)$ given $S=s$, denoted as $\nu_{f^* \mid s}$, has a density. Then, for the fair decision function space $\mathcal{F}$, we have
$$
\begin{aligned}
\underset{g\in \mathcal{F}} {\min}E\left|f^*(\boldsymbol{X},S)-g(\boldsymbol{X},S)\right|^2=\underset{\nu_g\in \mathcal{P}_2\left(\mathbb{R}\right) }{\min}\sum_{s \in \mathcal{S}} p_s \mathcal{W}_2^2\left(\nu_{f^* \mid s}, \nu_g\right),
\end{aligned}
$$
where $p_s=P(S=s)$ and $\nu_{g}$ denotes the distribution of $g(\boldsymbol{X}, S)$.
Moreover, if $g^*$ and $\nu^*$ are the solutions of the left-hand side and right-hand side minimization problems, respectively, then $\nu_{g^*}=\nu^*$ and
\begin{equation}\label{Gstar}
    g^*(\boldsymbol{x}, s)=\left(\sum_{s^{\prime} \in \mathcal{S}} p_{s^{\prime}} Q_{f^* \mid s^{\prime}}\right) \circ F_{f^* \mid s}\left(f^*(\boldsymbol{x}, s)\right),
\end{equation}
where the cumulative distribution function $F_{f \mid s}(t) = P(f(\boldsymbol{X}, S) \leq t \mid S=s)$ and the quantile function $Q_{f \mid s}(t) = \inf\{y \in \mathbb{R}: F_{f \mid s}(y) \geq t\}$.
\end{proposition}

Proposition \ref{thm2} demonstrates that any decision function derived from the existing methods \citep{qian2011performance,zhao2012estimating,zhang2012robust,liu2018augmented,fan2017concordance,Dlearning} can be directly transported to a fair version. Furthermore, the corresponding fair ITR (FITR) is $\operatorname{sgn}\left(g^*\right)$.


\subsection{Improved Trade-off ITR}
Although the fair decision function $g^*$ satisfies the fairness requirement, a typical challenge lies in balancing the maximization of value with the pursuit of fairness.
In real-world scenarios, it is crucial to prioritize relative fairness that meets specific fairness criteria,   rather than striving for absolute fairness.


To achieve different levels of fairness, we consider the disparate impact index as a metric for evaluating fairness. For any decision function $f$, we define the disparate impact of $f$ as:
$$
\text{DI}(f)=\min_{s,s^{\prime}\in \mathcal{S}}\frac{P(f(\boldsymbol{X}, S) >0 \mid S=s)}{P(f(\boldsymbol{X}, S) >0 \mid S=s^{\prime})}.
$$
If $\text{DI}(f)=1$, then the decision function $f(\boldsymbol{x}, s)$ is fair.
Furthermore, we require the decision function $f(\boldsymbol{x}, s)$ to satisfy $\text{DI}(f)\ge \rho, \rho \in (0,1)$. Specifically, when $\rho=0.8$, the corresponding ITR meets the 80\% Rule \citep{biddle2017adverse}, which is established by the U.S. Equal Employment Opportunity Commission. This rule serves as a critical benchmark for identifying potential discriminatory impacts in employment practices, including hiring, promotions, and other personnel decisions.

To achieve the specified level of fairness, such as the 80\% Rule, we consider propose an improved trade-off decision function that pose weights on optimal decision function $f^*$ and fair decision function $g^*$, that is
$$g_w(\boldsymbol{x},s)=w (\boldsymbol{x},s) f^*(\boldsymbol{x},s) + (1-w (\boldsymbol{x},s)) g^*(\boldsymbol{x},s),$$
where $w (\boldsymbol{x},s)\in[0,1]$ is the weight function.
By adjusting the value of the weight function, the decision function $g_w$ can achieve different levels of fairness. In particular, when $w (\boldsymbol{x},s)\equiv 0$, $\text{DI}(g_w)=1$.
However, considering fairness inevitably results in a loss of value, i.e. $\mathcal{V}(f^{*})-\mathcal{V}(g_w)$.
The following proposition  demonstrates that the loss of value is associated with the conditional average treatment effect $\tau^*(\boldsymbol{x}, s)=E\}R(1)-R(-1) \mid \boldsymbol{X}=\boldsymbol{x}, S=s\}$.
\begin{proposition}\label{lemm3.3}
For any decision function $f$,
the value loss is given by
\begin{equation}
\mathcal{V}(f^{*})-\mathcal{V}(f)=E\left\{|\tau^{*}(\boldsymbol{X}, S)| I((\boldsymbol{X},S) \in \mathcal{M}(f))\right\},
\end{equation}
where $\mathcal{M}(f)=\{(\boldsymbol{x},s)|f^*(\boldsymbol{x},s)f(\boldsymbol{x},s)\leq 0\}$.
\end{proposition}
Assuming $\mathcal{M}(g_\lambda)$ is not empty, Proposition \ref{lemm3.3} implies that a smaller $|\tau ^{*} (\boldsymbol{X}, S)|$ for $(\boldsymbol{X}, S) \in \mathcal{M}(g_\lambda)$ contributes to a smaller value loss, $\mathcal{V}(f^{*})-\mathcal{V}(f)$. Therefore, aiming to reduce the loss of value, we prioritize selecting the individuals $(\boldsymbol{x},s)$ with smaller $|\tau^{*} (\boldsymbol{x}, s)|$ into the set $\mathcal{M}(g_\lambda)$. We consider a self-adjusted weight function,
$$w^* (\boldsymbol{x}, s; \alpha)=1-\exp \left(-\alpha |\tau ^{*} (\boldsymbol{x}, s)|\right), $$
where $ w^* (\boldsymbol{x}, s;\alpha) \in [0, 1)$  increases as $\alpha$ ranges over $[0, \infty)$.
Subsequently, the corresponding decision function is $$g_\alpha(\boldsymbol{x},s)=w^* (\boldsymbol{x}, s;\alpha) f^*(\boldsymbol{x},s) + (1-w^* (\boldsymbol{x}, s;\alpha)) g^*(\boldsymbol{x},s),$$
which is referred to as the $\alpha$-level fair decision function.
Moreover, the corresponding ITR, $\text{sgn}(g_\alpha(\boldsymbol{x}, s))$,
is referred to as the $\alpha$-level fair ITR ($\alpha$-FITR).
As $\alpha$ varies from $0$ to $\infty$, the decision function $g_\alpha(\boldsymbol{x},s)$ transitions from the fair decision function $g^{*}(\boldsymbol{x},s)$ to the decision function $f^*(\boldsymbol{x},s)$ that maximizes the value.
We observe that  $\mathcal{M}(g_\alpha) \subseteq \mathcal{M}(g^*)$. For all $ (\boldsymbol{x},s) \in \mathcal{M}(g^*)$,
the individuals with smaller $|\tau ^{*} (\boldsymbol{x}, s)|$ are given a higher weight $1-w^* (\boldsymbol{x},  s;\alpha)$. This suggests that individuals with smaller $|\tau ^{*} (\boldsymbol{x}, s)|$
are prioritized for selection into
$\mathcal{M}(g_\alpha)$, resulting in a reduced value loss.


\subsection{Estimation Procedure }
Firstly, we discuss a general process for estimating a fair decision function based on any target dataset. Using the formulation of $g^*(\boldsymbol{x}, s)$ in (\ref{Gstar}), given the base estimator $\widehat{f}(\boldsymbol{x}, s)$ trained on a training dataset, we can derive a fair decision function $g^*(\boldsymbol{x}, s)$ using any other dataset. Consequently, our method operates as a plug-in algorithm. Assume that we have an independent and identically distributed (i.i.d.) training data set $\mathcal{L}_0$ containing treatment and reward information, represented as $\mathcal{L}_0 = \{(\boldsymbol{X}_i, S_i, A_i, R_i) \}_{i=1}^{n}$, drawn from the distribution $P$. Additionally, let $ \mathcal{L} = \{(\boldsymbol{X}_i, S_i)\}_{i=n+1}^{(n+N)} $ denote an i.i.d. target dataset for which we aim to achieve a fair treatment assignment.

 We partition the target dataset $\mathcal{L}$  into subsets based on sensitive attributes such that $\mathcal{L}=\cup_{s \in \mathcal{S}}\mathcal{G}^s$, where $\mathcal{G}^s=\{(\boldsymbol{X}_i,s), i \in \mathcal{I}^s \}$ represents the set of group-specific samples, and $\mathcal{I}^s=\{i \mid S_i=s,  i=n+1,\cdots,n+N\}$ denotes the indices of subsamples.
 The sample size in $\mathcal{G}^s$ is denoted as $N_s$. Consequently, the total number of samples in the target data set is given by $ N = \sum_{s \in \mathcal{S}} N_s $.




Let $\widehat{f}$ be an estimator of the optimal decision function, as described in the literature \citep{qian2011performance,zhao2012estimating,zhang2012robust,liu2018augmented,fan2017concordance,Dlearning}, which is estimated using the training data set $\mathcal{L}_0$.
Given the optimal decision function estimator $\widehat{f}$, we can empirically estimate the cumulative distribution function (CDF) $F_{f^* \mid s}$ and the quantile function $Q_{f^* \mid s}$ using the target dataset $\mathcal{L}$.
In the estimation procedure, we introduce the ``jittering" technique \citep{shen2020distance}, which involves adding a smoothing noise $\varepsilon$ to the estimator $\widehat{f}(\boldsymbol{x}, s)$ for tie-breaking. Specifically, we  consider the augmented dataset
$\{\widehat{f}\left(\boldsymbol{X}_i, s\right)+\varepsilon_{i}\}_{i \in \mathcal{I}^s}$, where the noise $\{\varepsilon_{i}\}_{i=n+1}^{n+N}$ are i.i.d. uniformly distributed on $[-\sigma, \sigma]$ and are independent of the datasets
$\mathcal{L}$ and $\mathcal{L}_0$. The parameter
$\sigma =O(n^{-\kappa_0/2})$ is chosen to match the convergence rate of $\widehat{f}$ specified in Assumption \ref{ASS2}.
Consequently, given the training data set $\mathcal{L}_0$, $\{\widehat{f}\left(\boldsymbol{X}_i, s\right)+\varepsilon_{i}\}_{i \in \mathcal{I}^s}$ consists of i.i.d. continuous random variables.

Based on the augmented dataset, the empirical CDF of $\widehat{f}(\boldsymbol{X}, s)+\varepsilon$ is given by:
$$\widehat{F}_{\widehat{f} \mid s}(t)=\frac{1}{N_s}  \sum_{i \in \mathcal{I}^s } I(\widehat{f}(\boldsymbol{X_i}, s)+\varepsilon_{i}<t). $$
Let $\widehat{Q}_{\widehat{f} \mid s}=\widehat{F}^{-1}_{\widehat{f} \mid s}(t)$ denote the empirical quantile function of $\widehat{f}(\boldsymbol{X}, s)+\varepsilon$ .
Furthermore, the empirical frequency is given by
$\widehat{p}_s=N_s/N$. Given a base estimator $\widehat{f}$ of the optimal decision function $f^*$ constructed from $\mathcal{L}_0$, we define the fair decision function estimator as 
\begin{equation}\label{GHAT}
\widehat{g}(\boldsymbol{x}, s)=\frac{1}{m} \sum_{l=1}^{m}\left[\left(\sum_{s^{\prime} \in \mathcal{S}} \widehat{p}_{s^{\prime}} \widehat{Q}_{\widehat{f} \mid s^{\prime}}\right) \circ\widehat{F}_{\widehat{f} \mid s}(\widehat{f}(\boldsymbol{x}, s)+\varepsilon_{l}^{\prime})\right],
\end{equation}
where $\{\varepsilon_{l}^{\prime}\}_{l=1}^{m}$ are i.i.d. samples, uniformly distributed on $[-\sigma, \sigma]$.
Using the two-step estimation method, which is referred as ``R-learner" \citep{nie2021quasi}, we obtain the CATE estimator $\widehat{\tau}$. Finally, the plug-in estimator of the proposed improved trade-off decision function is given by:
\begin{equation}\label{GHATLAMABA}
\widehat{g}_\alpha(\boldsymbol{x}, s)=\widehat{w }(\boldsymbol{x}, s;\alpha) \widehat{f}(\boldsymbol{x}, s)+(1-\widehat{w }(\boldsymbol{x}, s;\alpha)) \widehat{g}(\boldsymbol{x}, s),
\end{equation}
for any $\alpha \in [0, \infty)$, where $\widehat{w }(\boldsymbol{x}, s;\alpha)=1-\exp(-\alpha |\widehat{\tau}(\boldsymbol{x},s)|)$. An algorithm for the implementation of $\widehat{g}(\boldsymbol{x}, s)$ and $\widehat{g}_\alpha(\boldsymbol{x}, s)$ is provided in Algorithm \ref{alg:0}.

\begin{algorithm}[h]
    \caption{ Procedure to evaluate estimator $\widehat{g}(\boldsymbol{x}, s)$ and $\widehat{g}_\alpha(\boldsymbol{x}, s)$}
    \label{alg:0}
    \textbf{Input}: Training dataset $\mathcal{L}_0=\left\{\left(\boldsymbol{X}_i, S_i, A_i, R_i\right)\right\}_{i=1}^n$; target dataset  $\mathcal{L} = {(\boldsymbol{X}_i, S_i)}_{i=n+1}^{(n+N)}$; empirical frequencies $\widehat{p}_s=N_{s}/N$; group-specific index set $\mathcal{I}^s=\{i \mid S_i=s, i=n+1,\cdots,n+N\}$; the slack parameter $\sigma$; the number of jittering  $m$.\\
     \textbf{Output}: The fair decision function estimator $\widehat{g}(\boldsymbol{x}, s)$ and the $\alpha$-level fair decision function estimator $\widehat{g}_\alpha(\boldsymbol{x}, s)$ for the point $(\boldsymbol{x}, s)$.
    \begin{algorithmic}[1] 
        \STATE  Estimate the optimal decision function estimator $\widehat{f}$ based on $\mathcal{L}_0$.
        \STATE Estimate CATE $\widehat{\tau}(\boldsymbol{x},s)$ based on $\mathcal{L}_0$ via the R-learner method and obtain $\widehat{w }(\boldsymbol{x}, s;\alpha)$.
        \FOR{$s^{\prime} \in \mathcal{S}$}
        \STATE Obtain $\left\{\widehat{f}\left(\boldsymbol{X}_i, s^{\prime}\right)+\varepsilon_i\right\}_{i \in \mathcal{I}^{s^{\prime}} }$, where $\varepsilon_i\sim  U[-\sigma, \sigma]$.
        \STATE Estimate $\widehat{F}_{\widehat{f} \mid s^{\prime}}(t)$ and $\widehat{Q}_{\widehat{f} \mid s^{\prime}}(t)$.
        \ENDFOR
          \RETURN $\widehat{g}(\boldsymbol{x}, s)=\frac{1}{m} \sum_{l=1}^{m}\left[\left(\sum_{s^{\prime} \in \mathcal{S}} \widehat{p}_{s^{\prime}} \widehat{Q}_{\widehat{f} \mid s^{\prime}}\right) \circ\widehat{F}_{\widehat{f} \mid s}(\widehat{f}(\boldsymbol{x}, s)+\varepsilon_{l}^{\prime})\right], \forall \varepsilon_{l}^{\prime}\sim  U[-\sigma, \sigma]$.
          \RETURN $\widehat{g}_\alpha(\boldsymbol{x}, s)=\widehat{w }(\boldsymbol{x}, s;\alpha) \widehat{f}(\boldsymbol{x}, s)+(1-\widehat{w }(\boldsymbol{x}, s;\alpha)) \widehat{g}(\boldsymbol{x}, s),\forall \alpha \in [0, \infty)$.
    \end{algorithmic}
\end{algorithm}

\subsection{Computational Aspects}
In this section, our objective is to estimate the parameter $\alpha$ for $\widehat{g}_\alpha(\boldsymbol{x}, s)$, with the objective of achieving a specified fairness level regarding DI while simultaneously maximizing the value function $ \mathcal{V}(\widehat{g}_\alpha)$. To this end, we seek to solve the following optimization problem:
$$\max_{\alpha\in [0, \infty)} \mathcal{V}(\widehat{g}_\alpha)\text { subject to } \text{DI}(\widehat{g}_\alpha) \geq \rho, $$
where $\rho\in [0,1]$ is determined by the decision maker.

Following the value function proposed by \cite{zhao2012estimating}, we can reformulate the maximization problem as:
\begin{equation}\label{owl}
 \begin{aligned}\min_{\alpha\in [0, \infty)} E\left\{ \frac{RI(A \neq \operatorname{sgn}(\widehat{g}_\alpha\left(\boldsymbol{X}, S\right)))}{A \pi(1\mid\boldsymbol{X},S) +(1-A) / 2}\right\} \text { subject to } \text{DI}(\widehat{g}_\alpha) \geq \rho.
\end{aligned}
\end{equation}
The objective function involves a 0-1 loss, which is difficult to optimize due to its discontinuous and nondifferentiable nature. To address this issue, we employ a smoother surrogate $h(a,y)=-0.5(a+1) y+\log (1+y)$ as a replacement for the 0-1 loss \citep{jiang2019entropy}. This surrogate loss is continuous and differentiable.
{We replace Equation (\ref{owl}) as follows:}
\begin{equation}\label{entropyloss}
 \begin{aligned}
\min_{\alpha\in [0, \infty)} E\left\{\frac{Rh(A, \widehat{g}_\alpha\left(\boldsymbol{X}, S\right))}{A \pi(1\mid\boldsymbol{X},S)+(1-A) / 2}\right\}\text { subject to } \text{DI}(\widehat{g}_\alpha) \geq \rho.
\end{aligned}
\end{equation}
{According to the Fisher consistency \citep{jiang2019entropy}, the surrogate loss function
$h(a,y)$ used in Equation (\ref{entropyloss}) yields the same optimal solution as the 0-1 loss used in Equation (\ref{owl}).}

In addition, according to the definition of DI, the constraint condition is equivalent to $$ \min_{s,s^{\prime}\in \mathcal{S}}\rho{P(\widehat{g}_\alpha(\boldsymbol{X}, S) >0 \mid S=s^{\prime})}+{P(\widehat{g}_\alpha(\boldsymbol{X}, S) \leq 0 \mid S=s)}-1\leq 0,$$
that is,
$$\begin{aligned} \label{constraint}
&\min_{s,s^{\prime}\in \mathcal{S}}\frac{\rho}{P(S=s^{\prime})}E\{I(\widehat{g}_\alpha(\boldsymbol{X}, S)>0)I(S=s^{\prime})\} \\&+\frac{1}{P(S=s)}E\{I(\widehat{g}_\alpha(\boldsymbol{X}, S) \leq 0)I(S=s)\}-1\leq 0.
\end{aligned}$$
Accordingly, we consider the following optimization problem using the training dataset $\mathcal{L}_0$,
\begin{equation} \label{constraint}
  \begin{aligned}
 &\min _\alpha \frac{1}{n} \sum_{i=1}^n \frac{R_i}{A_i \widehat{\pi}(1\mid\boldsymbol{X}_i, S_i)+\left(1-A_i\right) / 2}h(A_i, \widehat{g}_\alpha\left(\boldsymbol{X}_i, S_i\right)) \\& \text { subject to }
\begin{aligned}
&\frac{\rho}{n_{s^{\prime}}} \sum_{i=1}^{n} I\left(\widehat{g}_\alpha\left(\boldsymbol{X}_i, S_i\right)> 0 \right)I(S_i=s^{\prime})+\frac{1}{n_s} \sum_{i=1}^{n} I\left(\widehat{g}_\alpha\left(\boldsymbol{X}_i, S_i\right)\leq 0 \right)I(S_i=s)\\
&-1  \le 0,  \forall s,s^{\prime} \in \mathcal{S},
\end{aligned}
\end{aligned}
\end{equation}
where $n_s=\sum_{i=1}^{n}I(S_i=s)$. {
Here, $\widehat{\pi}(a\mid\boldsymbol{x},s)$ is the estimated propensity score function, which is obtained using logistic regression.}

However, optimizing the indicator $I(\widehat{g}_\alpha(\boldsymbol{x}, s)>0)$ remains a challenge. Thus, we propose a smooth surrogate $H_{\beta,\gamma}(\widehat{g}_\alpha(\boldsymbol{x}, s))$ with $\beta,\gamma\in  \mathbb{R}^{+}$:
$$H_{\beta,\gamma}(u)=\left \{\begin{array}{cl}
\frac{1}{2}+\frac{1}{1+e^{-\frac{u}{\beta}}} & u \geqslant 0, \\\frac{1}{1-\gamma u} & u<0.
\end{array}\right.$$
In this formulation, the parameters $\beta$ and $\gamma$ are non-negative real numbers, which control the smoothness and asymptotic behavior of the surrogate function $H_{\beta, \gamma}(u)$ with respect to the indicator function $I(u>0)$.
Specifically, increasing $\beta$ and $\gamma$ can improve the approximation of $H_{\beta, \gamma}(u)$ to $I(u>0)$.
It should be noted that $H_{\beta, \gamma}(u)$ is almost everywhere differentiable, but it is not differentiable at $u=0$ unless the condition $\gamma=1/4 \beta$ is satisfied.

The smooth surrogate $H_{\beta,\gamma}(u)$ is a monotonically increasing function and satisfies $H_{\beta,\gamma}(u) \geq I({u\ge 0})$ for all $u \in {R}$. Hence, we have
\begin{equation}
\begin{aligned}
& \frac{\rho}{n_{s^{\prime}}} \sum_{i=1}^{n} I\left(\widehat{g}_\alpha\left(\boldsymbol{X}_i, S_i\right)> 0 \right)I(S_i=s^{\prime})+\frac{1}{n_s} \sum_{i=1}^{n} I\left(\widehat{g}_\alpha\left(\boldsymbol{X}_i, S_i\right)\leq 0 \right)I(S_i=s)-1\\
\leq &\frac{\rho}{n_{s^{\prime}}} \sum_{i=1}^{n} H_{\beta,\gamma}(\widehat{g}_\alpha(\boldsymbol{X}, S))I(S_i=s^{\prime})+\frac{1}{n_s} \sum_{i=1}^{n} H_{\beta,\gamma}(-\widehat{g}_\alpha(\boldsymbol{X}, S))I(S_i=s)-1.
\nonumber
\end{aligned}
\end{equation}
If the new constraint above is satisfied, then the original constraint in (\ref{constraint}) will also hold.
Finally,  the Equation (\ref{constraint}) can be substituted as follows:
\begin{equation}\label{optim}
 \begin{aligned}
 &\min _\alpha \frac{1}{n} \sum_{i=1}^n \frac{R_i}{A_i \widehat{\pi}(1\mid\boldsymbol{X},S)+\left(1-A_i\right) / 2}h(A_i, \widehat{g}_\alpha\left(\boldsymbol{X}_i, S_i\right)) \\& \text { subject to }
\begin{aligned}
&\frac{\rho}{n_{s^{\prime}}} \sum_{i=1}^{n} H_{\beta,\gamma}\left(\widehat{g}_\alpha\left(\boldsymbol{X}_i, S_i\right) \right)I(S_i=s^{\prime})+\frac{1}{n_s} \sum_{i=1}^{n} H_{\beta,\gamma}\left(-\widehat{g}_\alpha\left(\boldsymbol{X}_i, S_i\right) \right)I(S_i=s)\\
&-1  \le 0,  \forall s,s^{\prime} \in \mathcal{S}.
\end{aligned}
\end{aligned}
\end{equation}
To solve the constrained optimization problem (\ref{optim}), we recommend employing the Improved Stochastic Ranking Evolution Strategy (ISRES) \citep{runarsson2005search} and the pseudo-code is presented in Algorithm \ref{alg:1}. For the selection of the tuning parameters $\beta$ and $\gamma$, we use a grid search approach. Specifically, we define a grid of possible values for $\beta$ and $\gamma$: $\beta,\gamma \in \{100+10j | j \in \mathbb{Z}, 0\le j\le 90\}$. For each pair $(\beta,\gamma)$, we obtain $\widehat{\alpha}$ by solving the optimization problem (\ref{optim}), {and subsequently evaluate the performance of the resulting estimator $\widehat{g}_{\widehat{\alpha}}$ using empirical estimate of $\mathcal{V}(\widehat{g}_{\widehat{\alpha}})$,  defined as: $$\widehat{\mathcal{V}}(\widehat{g}_{\widehat{\alpha}})=\frac{1}{n} \sum_{i=1}^n \frac{R_iI(A =\operatorname{sgn}(\widehat{g}_{\widehat{\alpha}}\left(\boldsymbol{X}, S\right)))}{A_i \widehat{\pi}(1\mid\boldsymbol{X},S)+\left(1-A_i\right) / 2}.$$ Finally, the optimal hyperparameter pair is selected as the one that maximizes $\widehat{\mathcal{V}}(\widehat{g}_{\widehat{\alpha}})$.}
\begin{algorithm}[h]
    \caption{ Procedure to obtain the optimal solution $\widehat{\alpha}$ of optimization problem (\ref{optim})}
    \label{alg:1}
    \textbf{Input}: Training dataset $\mathcal{L}_0=\left\{\left(\boldsymbol{X}_i, S_i, A_i, R_i\right)\right\}_{i=1}^n$; group-specific index set $\mathcal{I}_{0}^s=\{i \mid S_i=s, i=1,\cdots,n \}$; the slack parameter $\sigma$; fairnes constraint $\rho\in [0,1]$.\\
     \textbf{Output}: Optimal solution $\widehat{\alpha}$ .
    \begin{algorithmic}[1] 
        \STATE  Estimate the optimal decision function estimator $\widehat{f}$ based $\mathcal{L}_0$.
        \STATE Estimate CATE $\widehat{\tau}(\boldsymbol{x},s)$ based $\mathcal{L}_0$ via the R-learner method and obtain $\widehat{w }(\boldsymbol{x}, s;\alpha)$.
        \FOR{$s\in \mathcal{S}$}
        \STATE Obtain $\left\{\widehat{f}\left(\boldsymbol{X}_i, s\right)+\varepsilon_i\right\}_{i \in \mathcal{I}_{0}^{s} }$, where $\varepsilon_i\sim  U[-\sigma, \sigma]$.
        \STATE Estimate $\widehat{F}_{\widehat{f} \mid s}(t)$ and $\widehat{Q}_{\widehat{f} \mid s}(t)$.
        \ENDFOR
        \FOR{$i\in\{1,\dots n\}$}
        \STATE Compute plug-in estimator $\widehat{g}_\alpha(\boldsymbol{X}_i, S_i)$
        for $\alpha \in [0, \infty)$ based on Equation (\ref{GHATLAMABA}).
        \ENDFOR
        \STATE Solving  (\ref{optim}) using the optimization algorithm ISRES.
          \RETURN  Optimal solution $\widehat{\alpha}$.
    \end{algorithmic}
\end{algorithm}

\section{Theoretical Results}
To establish the theoretical properties of the proposed estimators, we need the following conditions.
\begin{enumerate}[label=(A\arabic*), ref=(A\arabic*)]
    \item \label{cnd:1}  The outcome $R$ is bounded, with $|R|\le M$.
    \item \label{cnd:2} There exists some constant $M_0>0$ such that $E(g^*(\boldsymbol{X}, S))^2 \le M_0$.
    \item \label{cnd:3} For each $s \in \mathcal{S}$, the conditional distribution of $f^*(\boldsymbol{X}, S)\mid S=s$, denoted as $\nu_{f^* \mid s}$, has a density $q_s$, which has a lower bound $\underline{\lambda}_s>0$ and an upper bound $\bar{\lambda}_s \geq \underline{\lambda}_s$.
    \item \label{cnd:4} There exists some constant $\kappa_0>0$, such that $$E|f^*(\boldsymbol{X}, S)-\widehat{f}(\boldsymbol{X}, S)|^2=O(n^{-\kappa_0}).$$
    \item \label{cnd:5} There exists some constant $\kappa_{1}>0$, such that $$E|\widehat{\tau}\left(\boldsymbol{X},S\right)-\tau^*\left(\boldsymbol{X},S\right)|^2=O(n^{-\kappa_{1}}).$$
\end{enumerate}

{Conditions \ref{cnd:4} and \ref{cnd:5} require the estimated optimal decision function and CATE to satisfy certain convergence rates.
\cite{shi2020breaking} verified that Q-learning methods satisfy Conditions \ref{cnd:4} and \ref{cnd:5}.
Additionally, \cite{Dlearning} and \cite{fan2017concordance} showed that an optimal linear decision rule $\widehat{f}$ satisfies  Condition \ref{cnd:4}.  The estimated CATE $\widehat{\tau}(\boldsymbol{x},s)$, obtained from the $R$-learner proposed by \cite{nie2021quasi}, satisfies the convergence rate required by Condition \ref{cnd:5}.}

Proposition \ref{fairnguarantee} demonstrates a distribution-free fairness guarantee for the fair decision function estimator $\widehat{g}$.

\begin{proposition} \label{fairnguarantee}
Suppose  Conditions \ref{cnd:3} and \ref{cnd:4} hold and $\min_{s \in \mathcal{S}}N_s \rightarrow \infty$ when $N \rightarrow \infty$. Then the estimator $\widehat{g}$ defined in (\ref{GHAT}) satisfies
$$
 \lim_{n,N\to \infty}\max _{s,s^{\prime}\in \mathcal{S}}\sup _{t \in \mathbb{R}}\left|P(\widehat{g}(\boldsymbol{X}, S) \leq t \mid S=s)-P\left(\widehat{g}(\boldsymbol{X}, S) \leq t \mid S=s^{\prime}\right)\right|=0 .$$
\end{proposition}


\begin{theorem}[Convergence rate]\label{thmgAhat}
Using Conditions \ref{cnd:1}-\ref{cnd:5}, for any $\alpha \in [0, \infty)$, the estimator $\widehat{g}_\alpha$ defined in (\ref{GHATLAMABA}) satisfies
$$
E\left|g_{\alpha}(\boldsymbol{X}, S)-\widehat{g}_{\alpha}(\boldsymbol{X}, S)\right|
\le  O(n^{-\kappa_0/2})+ C\sum_{s \in \mathcal{S}} p_s N_s^{-1 / 2}+C\sqrt{\frac{|\mathcal{S}|}{N}} +\alpha O(n^{-\kappa_{1}/2}),
$$
where the constant $C= 2\sqrt{2\pi}
 \underline{\lambda}_0^{-1}+\sqrt{\pi/2} \left(\sum_{s \in \mathcal{S}} p_s \underline{\lambda}_s^{-1}\right)$ with $\underline{\lambda}_0=\min_{s \in \mathcal{S}}\underline{\lambda}_s$.
\end{theorem}
Consider the case where $p_s=1/|\mathcal{S}|$ and $N=n$, we obtain a simplified result:
$$E\left|g_{\alpha}(\boldsymbol{X}, S)-\widehat{g}_{\alpha}(\boldsymbol{X}, S)\right|=O(n^{-\min\{\frac{\kappa_0}{2}, \frac{1}{2}\}})+\alpha O(n^{-\frac{\kappa_{1} }{2}}).$$
The following theorem establishes the upper bound on the value loss of $\widehat{g}_\alpha$.
\begin{theorem}\label{thmvaluegAhat}
Under the conditions in Theorem \ref{thmgAhat}, for any $\alpha \in [0, \infty)$, the value loss of $\widehat{g}_\alpha$ satisfies
$$
\begin{aligned}
\mathcal{V}(f^*)-\mathcal{V}(\widehat{g}_\alpha) &\le  O(n^{-\kappa_0/2})+C_{1}\sum_{s \in \mathcal{S}} p_s N_s^{-1 / 2}+C_{1}\sqrt{|\mathcal{S}|} N^{-1 / 2}+\alpha O(n^{-\kappa_{1}/2})\\
&\quad+C_L \sqrt{\textstyle \sum_{s \in \mathcal{S}} p_s \mathcal{W}_2^2(\nu_{f^* \mid s}, \nu_{g^*})}\sqrt{E|\exp(-\alpha|\tau^*(\boldsymbol{X}, S)|)|^2},
\end{aligned}$$
where $C_1=CC_L$, $C_L=M/\tau$, and $C= 2\sqrt{2\pi}
 \underline{\lambda}_0^{-1}+\sqrt{\frac{\pi}{2}} \left(\sum_{s \in \mathcal{S}} p_s \underline{\lambda}_s^{-1}\right)$ with $\underline{\lambda}_0=\min_{s \in \mathcal{S}}\underline{\lambda}_s$.
\end{theorem}\label{the3}
Theorem \ref{thmvaluegAhat} indicates that the loss of value of $\widehat{g}$ is mainly determined by the Wasserstein distance between the optimal decision function $f^*$ and the fair decision function $g^*$.

\section{Simulation Studies}
\subsection{Simulation Designs}
We conducted  simulation studies to assess the performance of our proposed methods.  In simulations, we assumed that the outcome variable $R=R_0(\boldsymbol{X},S,A)+\epsilon$ where $\epsilon \sim N(0,1)$ represents the random error term. The function $R_0(\boldsymbol{X},S, A)$ reflects the effect of covariate $\boldsymbol{X}$, the sensitive attribute $S$, and the treatment variable $A$ on the outcome variable $R$.
The treatment variable $A$ was independently generated from $\{-1,1\}$, with $P(A=1)=1 / 2$. We generated 20-dimensional covariate vectors $\boldsymbol{X}=\left(X_1, \ldots, X_{20}\right)^{\top}$, where each component $X_i$ was independently drawn from $Unif(-5,5)$. Let sensitive attribute $S$ follow $Bernoulli(p_0)$, where the parameter $p_0$ is related to the covariate variables $\boldsymbol{X}$, defined as $p_0=$ $p\left(X_1\right) /\left(p\left(X_1\right)+p\left(X_2\right)\right)$ with $p(x)=\exp (x) /(1+\exp (x))$. We consider the following four experimental scenarios:\\
{\bf Experiment}  1:
$R_0(\boldsymbol{X},S, A)= 10+X_1+X_2+0.25 X_3 +\left(X_1+X_2-10(1-S)I(A=1)\right) A.$\\
{\bf Experiment}  2:
$R_0(\boldsymbol{X}, S,A)=10+\left(0.1X_1^2-X_2-10 (1-S) I(A=1)\right) A.$\\
{\bf Experiment}  3:
$R_0(\boldsymbol{X},S, A)=0.1
\left(-3-5X_1-X_2^2\right)
\left(5X_1+X_2^2-20-10S I(A=-1)\right) A+10.$\\
{\bf Experiment}  4:
$R_0(\boldsymbol{X},S,A)=\left(-0.5(X_1)^3+\log(X_2^2+1)+2 X_3 -0.5(X_4+X_5)^2-5\right)A+\left(10S I(A=1)\right)A+20.$

 In our simulation study, we considered the OWL method (introduced by \cite{zhao2012estimating}) to generate estimator $\widehat{f}$ of the optimal decision function.
 Specifically, we considered two types of decision functions: a linear function for Experiment 1, and nonlinear functions with Gaussian kernels for Experiments 2-4.
 For all four experiments, we generated training and testing datasets, each with a sample size of $500$.  Each simulation experiment was repeated $200$ times.
 We denoted the decision function learned by the OWL method, our proposed FITR and $\alpha$-FITR as $\widehat{f}_{OWL}$, $\widehat{g}_{FITR}$, and  $\widehat{g}_{\alpha}$, respectively. Specifically, we estimated the value function for each decision function as the empirical average of rewards obtained by executing the corresponding ITR on the testing dataset.
 In simulations, we estimated DI of $\widehat{g}_{\alpha}$ by calculating the following quantity on the testing dataset:
 $$ \widehat{\text{DI}}(\widehat{g}_{\alpha})=\min_{s,s^{\prime}\in \mathcal{S}}\frac{n_{s^{\prime}}\sum_{i=1}^{n} I\left(\widehat{g}_\alpha\left(\boldsymbol{X}_i, S_i\right)> 0 \right)I(S_i=s)}{n_{s}\sum_{i=1}^{n} I\left(\widehat{g}_\alpha\left(\boldsymbol{X}_i, S_i\right)>0 \right)I(S_i=s^{\prime})},$$
 where $s,s^{\prime} \in \{0,1\}$, $n_{s}$ and $n_{s^{\prime}}$
 denote the number of samples in the testing dataset with sensitive attribute $S=s$ and $S=s^{\prime}$, respectively.
Similarly, we denoted the estimated DI of $\widehat{f}_{OWL}$ and $\widehat{g}_{FITR}$ as $\widehat{\text{DI}}(\widehat{f}_{OWL})$ and $\widehat{\text{DI}}(\widehat{g}_{FITR})$.

\subsection{Simulation Results}
\begin{figure}[H]
	\centering
 	\begin{minipage}{0.45\linewidth}
		\centering
	\includegraphics[width=1\linewidth]{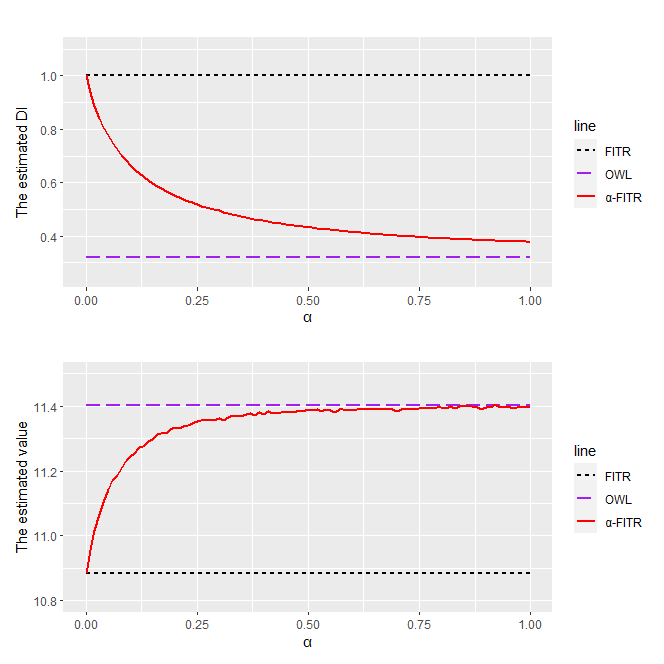}
        \subcaption{Experiment 1}
		\label{2a}
	\end{minipage}
        \hfill
  	\begin{minipage}{0.45\linewidth}
		\centering
	\includegraphics[width=1\linewidth]{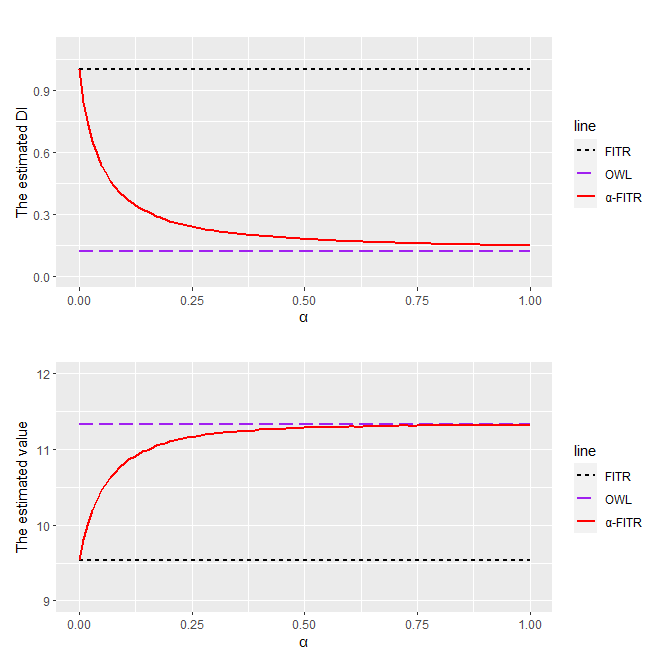}
        \subcaption{Experiment 2}
		\label{2b}
	\end{minipage}
 \hfill
 	\begin{minipage}{0.45\linewidth}
		\centering
	\includegraphics[width=1\linewidth]{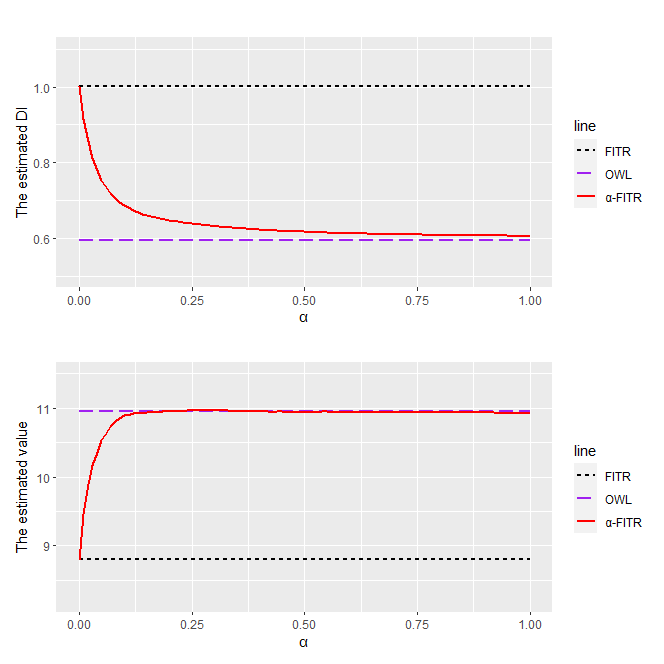}
        \subcaption{ Experiment 3}
		\label{2c}
	\end{minipage}
        \hfill
  	\begin{minipage}{0.45\linewidth}
		\centering
	\includegraphics[width=1\linewidth]{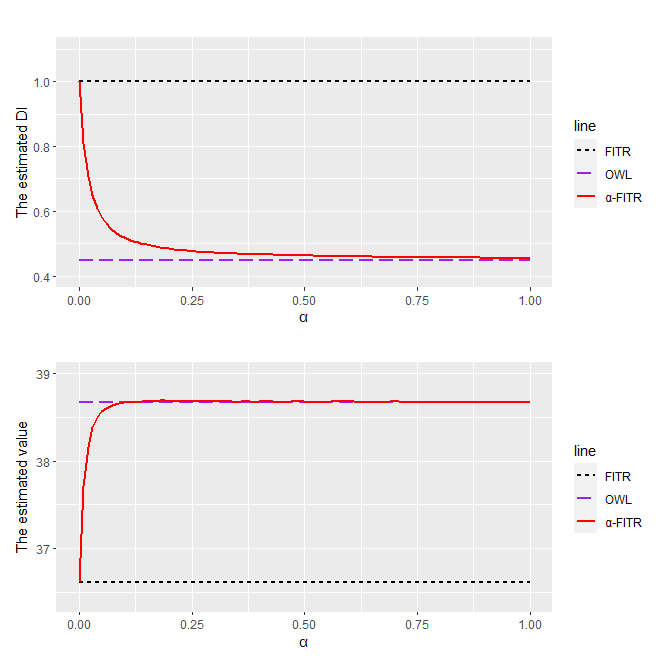}
        \subcaption{Experiment 4}
		\label{2d}
	\end{minipage}
         \caption{The estimated DI and value polts for the three ITRs: FITR, OWL, and $\alpha$-FITR, as $\alpha$ varies from 0 to 1 in increments of 0.01.}
         \label{fig2}
\end{figure}

Figure \ref{fig2} presents a comparison of value and fairness performance for  $\widehat{g}_{FITR}$, $\widehat{f}_{OWL}$, and  $\widehat{g}_{\alpha}$. In particular, for estimated DIs, our proposed fair ITR achieves $\widehat{\text{DI}}(\widehat{g}_{FITR})=1$ in all experiments, demonstrating its effectiveness in ensuring fairness in various scenarios.
The OWL method, i.e. $\widehat{f}_{OWL}$, exhibits certain drawbacks in terms of fairness. Specifically, as illustrated in Figures \ref{2a} and \ref{2b}, the estimated DIs of $\widehat{f}_{OWL}$ in Experiments 1 and 2 are consistently lower than $0.4$.

{Our proposed trade-off ITR exhibits a desirable trade-off between fairness and value. As shown in Figure \ref{fig2}, both the estimated DI and value of $\widehat{g}_{\alpha}$ (red line) exhibit a monotonic relationship with the parameter $\alpha$, characterized by a decreasing slope as $\alpha$ increases. Specifically, as $\alpha$ increases from $0$ to $1$, the level of the estimated DI decreases, indicating a reduction in fairness, while the estimated value of $\widehat{g}_{\alpha}$ increases. }

 Table \ref{tab1} displays the solutions $\widehat{\alpha}$ obtained by Algorithm \ref{alg:1} under different fairness requirements $\rho$ ($\text{DI}(\widehat{g}_{\alpha}) \geq \rho$), along with the DI estimates and value estimates for $\widehat{g}_{\widehat{\alpha}}$ based on the testing set. For all cases where $\rho<1.0$, the estimated DIs of $\widehat{g}_{\widehat{\alpha}}$ consistently larger than the corresponding values of $\rho$, providing validation for the optimization problem (\ref{optim}).

\begin{table}[h]
\centering
\caption{The estimated DIs (EDI) and estimated values (EV) of $\widehat{g}_{\widehat{\alpha}}$ ($\widehat{\alpha}$ solved by the Algorithm \ref{alg:1}) under different fairness requirements $\rho$ ($\text{DI} \geq \rho$) in Experiments 1,2,3 and 4.}
\scalebox{0.9}{
\begin{tabular}{@{\extracolsep{7pt}}cccccccc}
\toprule[2pt]
  & & $\rho=1.0 $ & $\rho=0.9$ &  $\rho=0.8$ & $\rho=0.7$ &  $\rho=0.6$&  $\rho=0.5$\\
\midrule
 \multirow{3}{*}{Experiment 1}  & $\widehat{\alpha}$ &0.000 & 0.020 & 0.051 & 0.101 & 0.180 & 0.335 \tabularnewline
 & EDI  & 0.999 & 0.902  & 0.806 & 0.708 & 0.619 & 0.537\tabularnewline
  & EV  & 10.890 & 11.014  & 11.107 & 11.205 & 11.276 & 11.316\tabularnewline \midrule
   \multirow{3}{*}{Experiment 2}   & $\widehat{\alpha}$ &  0.000 & 0.007 & 0.014 & 0.025 & 0.038 & 0.057 \tabularnewline
& EDI & 0.998 & 0.904  & 0.816 & 0.727 & 0.635 & 0.543\tabularnewline
& EV & 9.526 & 9.692  & 9.861 & 10.067 & 10.236 & 10.466\tabularnewline \midrule
 \multirow{3}{*}{Experiment 3}  &$\widehat{\alpha}$ &  0.001 & 0.015 & 0.048 & 0.158 & 0.413 & 0.729 \tabularnewline
& EDI & 0.992 & 0.912  & 0.827 & 0.742 & 0.668 & 0.622\tabularnewline
& EV & 8.770 & 9.168  & 9.782 & 10.215 & 10.595  & 10.765 \tabularnewline \midrule
 \multirow{3}{*}{Experiment 4}  &$\widehat{\alpha}$ &  0.001 & 0.003 & 0.008 & 0.019 & 0.059 & 0.200 \tabularnewline
& EDI & 0.981& 0.930  & 0.868 & 0.788 & 0.690 & 0.574\tabularnewline
& EV & 36.876 & 36.984  & 37.336& 37.850 & 38.214 & 38.689\tabularnewline \midrule
\bottomrule
\end{tabular}}
\par \label{tab1}
\end{table}

\section{Application}
We applied our proposed method to analyze real data from the Canadian entrepreneurship development program Next 36 \citep{lyons2017impact}.
The program, which started accepting applicants in 2011, aims to cultivate the next generation of innovators by providing them with practical education, mentorship, and access to a network of top entrepreneurs in the country.
For this analysis, we utilized a dataset derived from the Next 36 program sessions conducted between 2011 and 2015.
Specifically, the dataset consists of a total of 335 observations, comprising two distinct groups: 179 applicants who were accepted into the program ($A = 1$) and 156 applicants who were not accepted ($A = -1$).
The outcome variable $R$ is defined as follows: $R = 100$ if the individual engaged in any startup-related activity after the program ended (e.g., as a founder, co-founder, or employee), and $R = 0$ otherwise. We chose gender as the sensitive characteristic, with $89$ females ($S = 0$) and $246$ males ($S = 1$) in the data set, respectively.
 Two covariates were considered: the average score assigned by the interviewers and the school ranking. Based on this real dataset, our objective was to develop a fair admission decision framework tailored to applicants' characteristics while ensuring fairness with respect to gender. To conduct the analysis, we partitioned the data into two subsets: 50\% for training and 50\% for testing.

Specifically, using the OWL method, we utilized a Gaussian kernel to obtain estimator of optimal decision function, denoted by $\widehat{f}_{OWL}$. In this process, we estimated the propensity score $\pi(a\mid\boldsymbol{x},s)$ using the logistic regression. Using Algorithm \ref{alg:0}, we obtained estimators of fair decision function and $\alpha$-level fair decision function, denoted by $\widehat{g}_{FITR}$ and $\widehat{g}_{\alpha}$.

\begin{figure}[H]
\centering
     \includegraphics[width=1\textwidth]{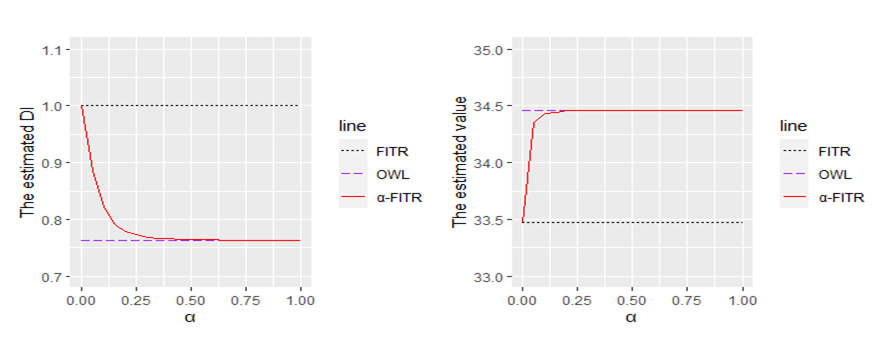}
\caption{The estimated DI and value polts for the three ITRs: FITR, OWL, and $\alpha$-FITR, as the conditioning parameter $\alpha$ varies from 0 to 1 in increments of 0.01.}
\label{figapply}
\end{figure}

As depicted in Figure \ref{figapply}, $\widehat{f}_{OWL}$ does not meet the 80\% rule \citep{biddle2017adverse}, as the estimated DI does not attain $0.8$, which indicates that the proportion of females accepted into the program is less than 80\% of the proportion of males.
This suggests a significant disparity in admission rates between the two genders. The proposed $\widehat{g}_{FITR}$ achieves the estimated DI that approaches $1$,
demonstrating its effectiveness in correcting the disparity in admission rates between males and females.

In Figure \ref{figapply}, the estimated DI and value of $\widehat{g}_{\alpha}$ exhibit a sharp change when $\alpha$ falls within the range $[0,0.125]$, followed by a relatively flat slope within $[0.125,0.25]$. {By adjusting $\alpha$ in $[0.125,0.25]$,} we can easily achieve $\text{DI}(\widehat{g}_{FITR})$ larger than $0.8$, with a slight decrease in value, making it a promising approach for real-world applications.

\section{Discussion}
We proposed a post-processing fairness framework for estimating fair ITRs that fully conform to demographic parity requirements. The proposed framework is a generic method that can be applied to any optimal decision function obtained by existing approaches, offering improved computational efficiency compared to methods that directly incorporate fairness constraints in the ITR estimation.
 Furthermore, we introduced a trade-off ITR that employs a self-adjusted weighting function to balance fairness and value optimization, thereby enhancing the fairness level of ITRs and ensuring compliance with legal standards of fairness.

Our post-processing fairness approach can be generalized to multiple treatment settings. For multiple treatments, the MOML method \citep{zhang2020multicategory} estimates the decision function $\boldsymbol{f}$ (($k-1$)-dimensional classification function vector) by  the weighted angle-based method. Building on the MOML method, we can consider transforming the estimated decision function $\widehat{\boldsymbol{f}}$ into a fair version, and then a similar generalization of our proposed method may be possible for fair decision-making in multiple treatment scenarios.

Exploring the applicability of our framework in dynamic regimes is another direction. For example, using the entropy learning method of \cite{jiang2019entropy}, which iteratively optimizes decision rules between stages, our post-processing fairness approach could be integrated at each stage to ensure fairness.

\newpage
\begin{center}
{\large\bf SUPPLEMENTARY MATERIAL}
\end{center}
The supplementary material includes all theoretical proofs.

\bibliographystyle{chicago}
\bibliography{Bibliography-MM-MC}

\end{document}